%% file: iclr2022_conference.tex
\documentclass{article} 
\usepackage{iclr2022_conference,times}

\input{math_commands.tex}

\usepackage{hyperref}
\usepackage{url}

\usepackage{subcaption} 
\usepackage{adjustbox}
\usepackage{multicol}
\usepackage{multirow}
\usepackage{amsmath}
\usepackage{booktabs}

\title{Measure Twice, Cut Once: \\ \large Quantifying Bias and Fairness in Deep Neural Networks}

\author{Cody Blakeney, Gentry Atkinson, Nathaniel Huish, Vangelis Metris, \& Ziliang Zong \\
Department of Computer Science\\
Texas State University\\
San Marcos, TX 78758, USA \\
\texttt{\{cjb92,gma23,njh71,vmetsis,ziliang\}@txstate.edu} \\
\And
Yan Yan \\
Department of Computer Science\\
Illinois Institute of Technology\\ 
Chicago, IL 60616, USA \\
\texttt{yyan34@iit.edu} \\
}

\iclrfinalcopy 
\begin{document}

\maketitle

\begin{abstract}
Algorithmic bias is of increasing concern, both to the research community, and society at large. Bias in AI is more abstract and unintuitive than traditional forms of discrimination and can be more difficult to detect and mitigate. A clear gap exists in the current literature on evaluating the relative bias in the performance of multi-class classifiers. In this work, we propose two simple yet effective metrics, Combined Error Variance (CEV) and Symmetric Distance Error (SDE), to quantitatively evaluate the class-wise bias of two models in comparison to one another. By evaluating the performance of these new metrics and by demonstrating their practical application, we show that they can be used to measure fairness as well as bias. These demonstrations show that our metrics can address specific needs for measuring bias in multi-class classification. 
\end{abstract}

\section{Introduction}

Broad acceptance of the large-scale deployment of AI and neural networks depends on the models' perceived trustworthiness and fairness. However, research on evaluating and mitigating bias for neural networks in general and compressed neural networks in particular is still in its infancy. Because deep neural networks (DNNs) are "black box" learners, it can be difficult to understand what correlations they have learned from their training data, and how that affects the downstream decisions that are made in the real world. Two models may appear to have very similar performance when only measured in terms of accuracy, precision, etc. but deeper analysis can show uneven performance across many classes. Moreover, when the number of tasks grows beyond one or two, the difficulty in reasoning and quantifying trade-offs when selecting or validating a model also grows.

\begin{figure}[h!]
     \centering

     \begin{subfigure}[b]{0.49\textwidth}
         \centering
         \includegraphics[width=\textwidth]{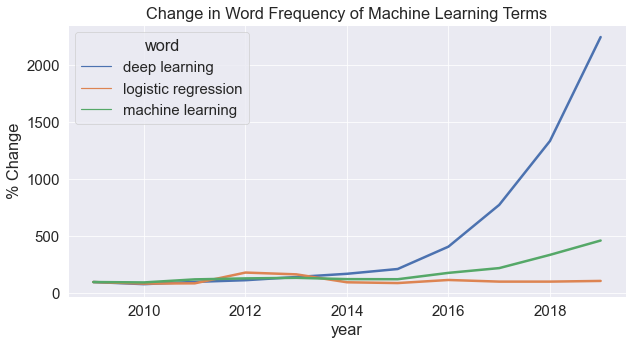}
        \caption{Deep Learning vs Logistic Regression}
        \label{fig:DLvLR}
     \end{subfigure}
     \begin{subfigure}[b]{0.49\textwidth}
         \centering
         \includegraphics[width=\textwidth]{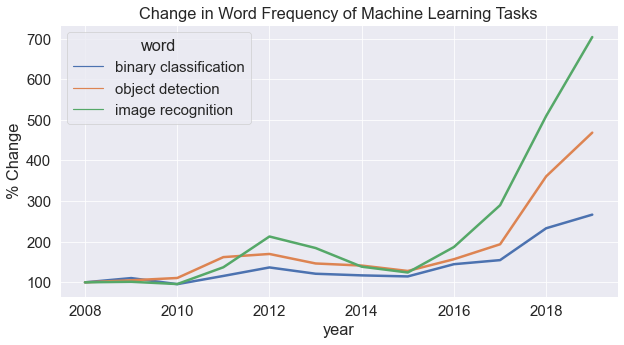}
        \caption{Change in Task Word Frequency}
        \label{fig:OD,IR,BC}
     \end{subfigure}

    \caption{Google NGram Data \cite{michel2011quantitative} showing relative usage of machine learning related terms over time. Deep learning has quickly passed up the use of statistical terms like logistic regression. Computer Vision tasks like Object Detection and Image Recognition are growing at faster rates than Binary Classification which fairness metrics can address. 
    }
    \label{fig:ML_trends}
    
\end{figure}

Widely accepted and effective metrics for measuring the bias of several neural networks against one another are still missing. Issues of both fairness and bias, which will be discussed as distinct but related phenomena in this paper, can seriously degrade the trustworthiness of a machine learning model in real-world conditions. It is important to quantify the performance of models in terms both of bias and fairness. While there exists extensive work on AI fairness regarding binary classification tasks~\cite{borkan2019nuanced, hinnefeld2018evaluating, dixon2018measuring, maughan2020towards}, there is a shortage of metrics extending these ideas to other machine learning tasks. Many researchers have taken an interest in wanting to ensure their models are fair, currently, we simply do not have the tools to measure for many domains.

In fact, recent trends shown in Figure \ref{fig:ML_trends}, are exacerbating the divide with the majority of new research in neural networks exploring topics outside of the reach of existing fairness and bias metrics. While difficult to quantify exactly we see from Google NGram data \cite{michel2011quantitative}  since 2010 the focus of machine learning is increasingly on multi-class classification and other difficult to quantify tasks rather than binary classification, revealing a need for metrics that can accommodate multi-class tasks. Worse still, we appear to be at the edge of another inflection point in AI where Large Language Models (LLMs) and Foundational models \cite{bommasani2021opportunities} like GPT-3 \cite{brown2020language} are ingesting the entire  corpus of human thought with limited supervision.


In this paper, we introduce two new metrics based on simple principles, whose purpose is to quantify a change in per-class bias between two or more models. These metrics provide singular data points that are easier to consider than the laborious checking of distributions of class-wise error rates. We will discuss their intuition and their application for comparing the relative performance of deep learning models, and classifiers in general, in terms of bias and fairness. To the best of our knowledge, these new metrics are distinct from all existing methods in that they expose per-group, per-class bias not neatly captured by other metrics, enabling the examination of issues of fairness and bias in great depth. While our proposed work is not a panacea for all emerging trends in AI we believe it represents a starting point to address the current gap.



The remaining sections are organized as follows. In Section~\ref{sec:background} we will contextualize the field of fairness metrics and their shortcomings as they relate to our considered problem domain. In Section~\ref{sec:metrics} we will define the intuition for our metrics, and provide a mathematical definition. In Section~\ref{sec:cases} we will provide specific use cases as experiments we envision the metrics will be used in, and how to reason about their differences. Section~\ref{sec:discuss} will discuss some limitations of our metrics and how we might improve or extend them to other domains.

\section{Background}
\label{sec:background}

Bias and fairness in machine learning have received increasing attention in recent years. The advantages of algorithmic decision-making can be very attractive to large organizations, but there is a risk that the output of these algorithms can be unfair~\cite{mehrabi2019survey}. Unfairness can have serious perceptual and legal consequences for organizations who choose to rely on machines to make important decisions~\cite{caton2020fairness}. This makes it imperative that quantitative measures for bias and fairness in machine learning be defined.

Bias, discrimination, and unfairness are terms that are often used interchangeably but we would like to make a distinction to better dissect the problem. We will refer to bias as meaning the behavior of a machine learning model giving preference to one characterization over another~\cite{mooney1996comparative}. Or put simply, having a lower error rate on one class than another. When discussing fairness in this paper, we will be referring to group fairness, which in general is concerned with outcomes for privileged and unprivileged groups~\cite{maughan2020towards}, where a group is a protected feature of an instance from the training data characterizing the instance in some way. Typically we do not want membership in a group to affect the outcome of a prediction, e.g. considering race or gender for ranking resumes or home loan applications.

There are other accepted definitions of fairness as well~\cite{mehrabi2019survey}. Individual fairness, requiring a model given similar predictions for similar individuals. Subgroup fairness, which uses notions of both individual and group fairness by holding some constraint  over large collections of a subgroup. However, group fairness is the most commonly measured by metrics of fairness~\cite{mehrabi2019survey}. 

Both bias and unfairness can degrade the performance of a model in ways that are not well captured by accuracy, precision, and other measures of ML performance. Biased and unfair models can perform very well on biased or unfair data. A nuanced metric can reveal conditions under which a model's performance might be degraded by the bias of the model or its training data. Many good metrics exist for measuring the group or individual unfairness of a model, but the focus has overwhelmingly been on tasks of supervised, binary classification~\cite{caton2020fairness}.

Substantial literature has emerged concerning algorithmic bias, discrimination, and fairness. Mehrabi et al. conducted a survey on bias and fairness in machine learning~\cite{mehrabi2019survey}. Mitchell et al. explored how model cards can be used to provide details on model performance across cultural, racial, and inter-sectional groups and to inform when their usage is not well-suited ~\cite{mitchell2019model}. Gebru el al. proposed using datasheets as a standard process for documenting datasets~\cite{gebru2018datasheets}. Amini,et al. proposed to mitigate algorithmic bias through re-sampling datasets by learning latent features of images~\cite{amini2019uncovering}. Wang et al. designed a visual recognition benchmark for studying bias mitigation in visual recognition.~\cite{wang2020towards}.

Other metrics of fairness have been described in recent works~\cite{borkan2019nuanced, hinnefeld2018evaluating, dixon2018measuring, maughan2020towards} whose purpose is to measure unfairness in machine learning models. Measurements of fairness based on the area under the receiver operating characteristic curve~(AUC-ROC) are described in~\cite{dixon2018measuring} and expanded in~\cite{borkan2019nuanced}. These metrics measure group-wise accuracy using AUC. Prediction Sensitivity, described in~\cite{maughan2020towards} fills the need for a reliable measure of individual fairness, as opposed to group fairness. A common shortcoming of these metrics is that they focus exclusively on binary classification~\cite{caton2020fairness} and are not meaningful in tasks of multi-class classification. Our proposed metrics are usable with any number of classes. Additionally, few works have studied how bias can present as unfairness and vice versa. To the best of our knowledge, our work is the first to propose a single metric shown to express both bias and unfairness when comparing two models.

\section{Proposed Metrics for Class-Wise Bias}
\label{sec:metrics}

We propose two new metrics, Combined Error Variance (CEV) and Symmetric Distance Error (SDE)s. Both measure changes in the class-wise false positive and false negative rates of two models, and each has its own advantages which will be explored in Section \ref{sec:cases}. When calculating both CEV and SDE one model is used as the base and another model as the alternative.

\subsection{Combined Error Variance}

The concept of the Combined Error Variance (CEV) metric is to measure the tendency of DNNs to sacrifice one class for the benefit of another class. CEV approximates the variance of the change in False Negative Rate~(FNR) and change in False Positive Rate~(FPR). It summarizes changes in FNR/FPR away from the model's average. Mathematically, CEV is defined as follows.

\begin{multicols}{2}
    \noindent
    \begin{equation} \label{norm_change}
        \delta{X_{ie}} =  \frac{X_{ie} - \hat{X_{ie}}}{\hat{X_{ie}}}
    \end{equation}
    \begin{equation}\label{mean_change}
        \delta{X}_{\mu e} = \frac{1}{n}\sum_{i=0}^{n} (\delta{X_{ie}})
    \end{equation}
\end{multicols}
\noindent
\begin{equation} \label{cev}
    cev = \frac{1}{n}\sum_{i=1}^{n} (dist((\delta{X}_{\mu pos}, \delta{X}_{\mu neg}), (\delta{X_{i pos}}, \delta{X_{i neg}})))^2
\end{equation}

Let $X_{ie}$ be a pair of values for the FPR and FNR for class $i$ of the comparison model and $\hat{X_{ie}}$ be the original models FPR/FNR pair, with e indicating either the false-positive or false-negative rate. We first find the normalized change in FPR/FNR $\delta{X_{ie}}$ by subtracting the error rates for the two models from each other and dividing by the original. The mean change $\delta{X_{\mu e}}$ is found by averaging the values of $\delta{X_{ie}}$, keeping in mind that every $\delta{X_{ie}}$ is the change in two values FPR and FNR. The CEV is calculated by treating each $\delta{X_{ie}}$ as a point in a 2-dimensional space of FNR and FPR. The square of the euclidean distances between each $\delta{X_{ie}}$ and the mean change represented by $\delta{X_{\mu e}}$ are summed and divided by the total number of classes $n$. 

\subsection{Symmetric Distance Error}

The principle of the Symmetric Distance Error (SDE) metric is to measure another undesirable bias behavior that presents in simple models. That is, a class with more training examples or that has similar features to another class is more frequently to be chosen by the model with limited capacity. To reflect this biased behavior, SDE calculates how "far away" from balanced is the change in FPR/FNR for each single class error. Intuitively, if we make a scatter plot with changes in FPR and FNR as X and Y values, the diagonal line in that plot would be a perfectly balanced change in FPR/FNR. Therefore, the SDE can be calculated as the symmetric distance of each change to that balance line.  

\begin{multicols}{2}
    \noindent
    \begin{equation} \label{SDE_base}
        d = \frac{|a(x_0) + b(y_0)|}{\sqrt{a^2 + b^2}}
    \end{equation}
    \begin{equation}
        d = \frac{|(1)(x_0) + (-1)(y_0)|}{\sqrt{(1)^2 + (-1)^2}}  = \frac{|x_0 -y_0|}{\sqrt{2}}
    \end{equation}
\end{multicols}

For a line in the Cartesian plane described by the equation $ax + by + c = 0$, the distance $d$ from any point $(x_0, y_0)$ can be derived from the equation in \ref{SDE_base}. In our specific context, the diagonal of the Cartesian plane (i.e. the balance line) is $x = y$ or $x - y = 0$ will represent an equal difference in FNR and FPR between two models. Given any change of FNR and FPR the symmetric distance of that change to the balance line can be calculated as:  

\noindent
\begin{equation}\label{sym_error}
sde = \frac{1}{n}\sum_{i=0}^{n}| \delta{FNR_i} - \delta{FPR_i} |
\end{equation}

Once the symmetric distance of each change is calculated, the SDE of a model can be calculated as the mean absolute change of normalized FP/FN rate, with the change being calculated as described in Equation~\ref{norm_change}. The $\sqrt{2}$ has been omitted from the final equation as a constant that has no effect on the meaning of the metric.  This metric will therefore reveal that one model or the other is more biased toward false positives or false negatives in a class-wise fashion.

\subsection{Normalization}
It is frequently true that a metric is meaningless without some numerical context. Both CEV and SDE may produce a large range of values depending on the specific dataset, number of classes, and performance of the models trained on that data. While not strictly necessary, in order to make the outputs of our metrics more interpretable, we follow a procedure for normalizing their values based on a hypothetical ''worst performing" model to give us a reference. To do this a set of predictions for all test instances is produced at random with all classes being equally likely, and the FPR/FNR of these random predictions is calculated. The CEV and SDE of the random predictions is generated relative to the original model. These are then used as a divisor to normalize the other CEV and SDE values of a group of models. Following this process, our metrics now indicate a change in algorithmic bias relative to a random predictor. Thus, a CEV value of 0.5 shows that the class-wise bias of model 2 relative to model 1 has increased by 50\% of the change between model 1 and a random predictor.

\section{Example Applications}
\label{sec:cases}

We have explored several applications of CEV and SDE for comparing the performance of two models. While we don't believe this list is exhaustive, in this section we illustrate several ways our proposed metrics can be used. We group these applications into two categories: 

\begin{enumerate}
    \item Comparing models w.r.t each other for the purpose of evaluating change in bias. We demonstrate using CEV/SDE in the context of model compression to detect compression-induced bias. We then further generalize this concept by informing and selecting from any number of trained low resource models to replace a higher capacity model.
    
    \item Evaluating group fairness. We demonstrate how CEV/SDE can be used to measure relative bias w.r.t protected groups. We also compare our results to existing binary classification fairness metrics and demonstrate the use of our metrics on multi-class data.
\end{enumerate}

\subsection{Model Selection}

\subsubsection{Model Compression}
Compressed neural networks can offer significant reductions in the computing power required for model inference. However, it has been shown that compressed models are often more biased than the original when the per-class error rates are examined~\cite{hooker2019compressed, hooker2020characterising}. CEV and SDE allow one to reason about two values instead of auditing all FP/FN rates of the classes. They can quickly expose compressed models that cannibalize a subset of their target classes to preserve top-1/top5 accuracy.

\subsubsection*{Compression Identified Exemplars}

To the best of our knowledge, only one other work proposes a method for measuring desperate impact and bias for multi-class classification. Compression Identified Exemplars (CIEs) ~\cite{hooker2019compressed, hooker2020characterising} are proposed specifically to categorize changes in model behavior and attribute the extent to which model compression techniques are responsible. While CIEs have many advantages beyond measuring bias (i.e. human-in-the-loop data auditing) the means by which they are found make counting CIEs impractical for many problems and impossible for many others. CIEs are defined as the following equation, which represents images in a data set for which compression explicitly changes the behavior of the answer. In~\cite{hooker2019compressed},  a population of 30 models were trained for each compression method and sparsity level. They then defined an image $i$ as an exemplar if modal label $Y_{i,t}^{M}$, or class most predicted by the \textit{t}-compressed model population disagrees with the label produced from the original networks. 

\[
CIE_{i,t} = \begin{cases}
1,  & \text{if } y_{i, 0}^M \neq y_{i, t}^M\\
0,  & \text{otherwise}
\end{cases}\
\]

While CIEs help reveal the bias issues in pruned models, not every image reported as a CIE represents a problem. A good portion of CIEs represent images that are equally hard for a human to classify and may simply be a case where the uncompressed networks overfit to learn the example. Our proposed metrics, CEV SDE, address some of the weaknesses of CIEs and give us true metrics. Pruning and quantization have been observed to sacrifice accuracy on a subset of classes in classification tasks in order to retain overall top-k accuracy~\cite{hooker2019compressed}. To catch and reflect this bias, our metrics are designed to quantify both the \textit{spread} of the change in classification error as well as changes in \textit{how} the model is making mistakes. As a result, both of our proposed metrics consider the distribution of change in false positive and false negative rates (FPR, FNR) for all classes.



To demonstrate the efficacy of CEV and SDE for evaluating bias in compressed models, we evaluate the following well-known compression algorithms and compare CEV and SDE metrics with the CIEs counts. The following example is conducted to illustrate the intuitive application of CEV/SDE. In this experiment, we measure the change that structured pruning has on a convolutional neural network as well as how various distillation methods mitigate bias. We train a CNN ResNet~\cite{he2016deep} (ResNet32x4) model on CIFAR100~\cite{Krizhevsky09learningmultiple} and prune it to various sparsities using Filter-wise Structured Pruning.

Hooker et al. reported that the negative effects of compression are most observed on underrepresented groups~\cite{hooker2020characterising}. Therefore, we resample the CIFAR100 dataset to purposely underrepresent certain classes in the training set. We select 15 classes and reduce the remaining training samples to 50\%, 20\%, and 10\% of their original numbers. Our hypothesis is that these biased classes with fewer data samples will more likely be picked as "victims" by the pruned models.  

\begin{figure}[t!]
     \centering
     \begin{subfigure}[b]{0.45\textwidth}
         \centering
         \includegraphics[width=\textwidth]{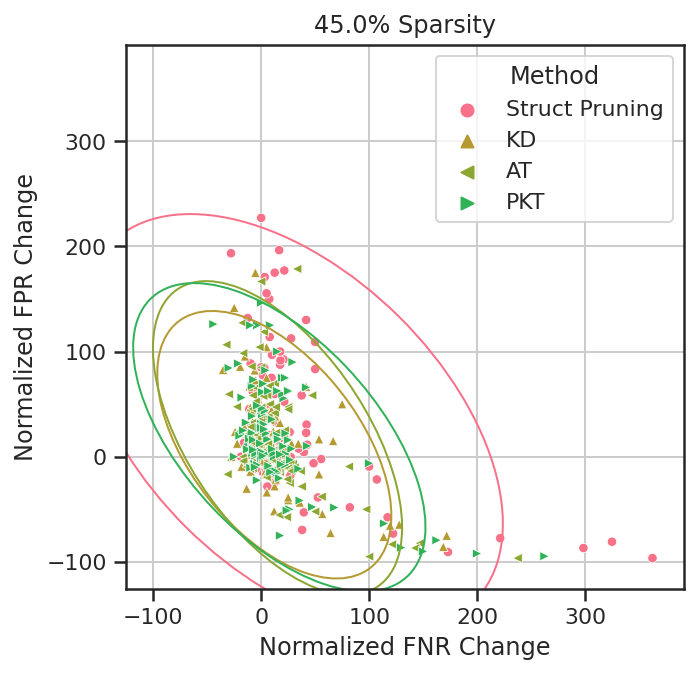}
        \caption{Pruning vs Distillation}
        \label{fig:PvFSP}
     \end{subfigure}
     \begin{subfigure}[b]{0.45\textwidth}
         \centering
         \includegraphics[width=\textwidth]{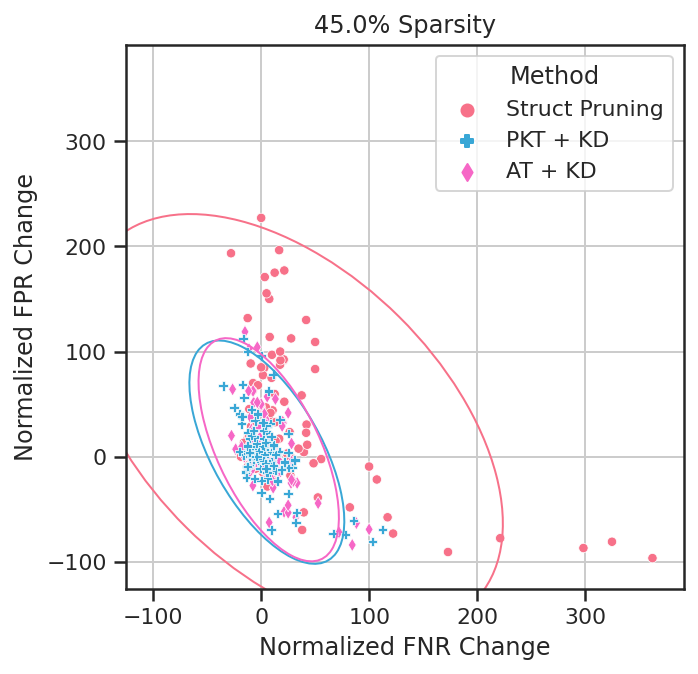}
        \caption{AT/PKT + KD}
        \label{fig:Sing_v_Mul}
     \end{subfigure}

    \caption{Normalized FP/FN Rate Change for  distillation methods on biased CIFAR100 dataset. Ellipses represent 95\% confidence intervals for data}
    \label{fig:bias}
    
\end{figure}



We use pruning as the baseline model and the rest of the models are pruned jointly with one of the state-of-the-art KD methods shown in Table~\ref{tab:similarity_tb}. We also test whether combining our feature map based distillation methods with KD (e.g. AT + KD or PKT + KD) can achieve additional benefits. We utilize Tian et al's~\cite{tian2019contrastive} implementation for all distillation methods tested. For all experiments, the models are pruned initially to 10\% sparsity then gradually pruned every 5 epochs until the desired sparsity is reached according to the AGP~\cite{zhu2017prune} schedule until they reach 45\% sparsity. All pruning is completed at the halfway point of training and allowed to continue to fine-tune for another 120 epochs. We do not perform any layer sensitivity analysis or prune layers at different ratios. Although that may have resulted in higher accuracy, our goal is not to reach state-of-the-art compression ratios but to demonstrate how CEV/SDE capture the effect of pruning on bias and any methods that might mitigate it.

\begin{table}[!h]
\centering
\caption{CIE count, CEV, SDE, and Accuracy for pruning with and without KD on biased CIFAR100}
\resizebox{.95\columnwidth}{!}{

\begin{tabular}{lrrrr}
\toprule
             Method & \# of CIEs &  CEV &  SDE &  Accuracy \\
\midrule
        AT + KD & 742    &               0.00187 &                    0.13173 &    77.100 \\
      PKT + KD & 748    &                0.00199 &                    0.13098 &    77.335 \\
        SP + KD & 768    &               0.00331 &                    0.16162 &    76.927 \\
      FSP + KD & 742    &                0.00333 &                    0.16002 &    75.285 \\
             KD \citep{hinton2015distilling} & 770    &               0.00338 &                    0.16065 &    78.142 \\
             AT \citep{Zagoruyko2017AT} & 909    &               0.00430 &                    0.19306 &    78.097 \\
            PKT \citep{pkt_eccv}  & 881    &               0.00481 &                    0.19891 &    78.963 \\
             SP \citep{tung2019similarity}  & 838    &               0.00583 &                    0.21591 &    78.520 \\
            FSP \citep{yim2017gift} & 877    &               0.00638 &                    0.22525
 &    78.413 \\
 Struct Pruning & 887    &               0.00931 &                    0.26687 &    77.242 \\
\bottomrule
\end{tabular}
}
\label{tab:similarity_tb}
\end{table}

Two observations can be made from the results presented in Table~\ref{tab:similarity_tb} and Figure~\ref{fig:PvFSP}: (1) CEV and SDE generally agree with the CIE count. They achieve this while being easier to calculate and not requiring multiple models to be trained. (2) Accuracy alone is a poor indicator of model quality. In Table~\ref{tab:similarity_tb} Structured pruning has accuracy comparable to AT + KD and PKT + KD but in Figure~\ref{fig:bias} we see that the change in accuracy in structured pruning is not distributed equitably with some classes having a 300\% change in FNR. SDE also neatly captures that the skew in FP/FN resulting for most models. These experiments show the great potential of our proposed CEV/SDE metrics in distinguishing desirable models from biased models that appear to be equal at a surface level.

\subsubsection{Pretrained Model Selection}

\begin{figure*}[h!]
    \centering
    \includegraphics[width=.9\textwidth]{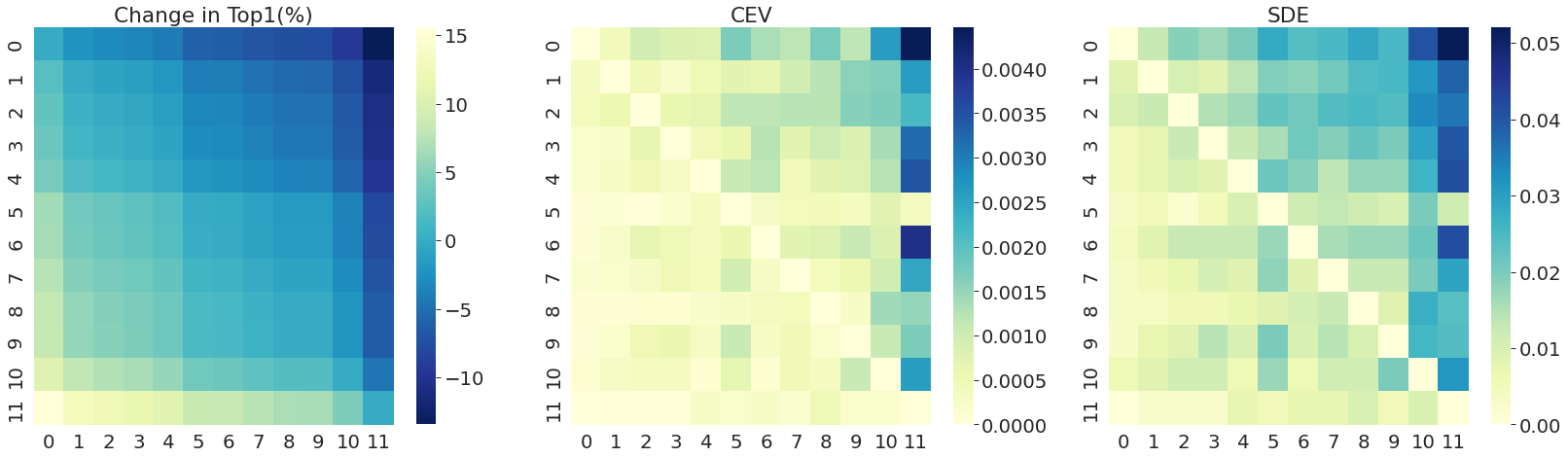}
    \caption{Change-in-Top1, normalized CEV, and normalized SDE adjacency matrices of models listed in Table \ref{tab:smallest_models}. Each entry displays the given metric calculated for the columns model w.r.t the rows model. Reading the table row-wise you see trade offs going from the row model to another. Reading the column you see the trade offs for other models going to the column model. Higher values CEV/SDE (shown by darker cells) indicate moving towards a more biased model.}
    \label{fig:small_models}
\end{figure*}

There are many algorithms and model architectures for low resource inference in image recognition alone. We have discussed examples of using CEV/SDE to analyze compression effects. Now we consider this problem more broadly. Bias is an important consideration when selecting a pre-trained model from one of the dozens which are available in many problem spaces. Unfortunately, CIE count is not applicable in the case where you have models already trained and simply want to understand the trade-offs you will be making. Here we see how one might use CEV/SDE to detect and avoid a model more biased than models of similar accuracy.
\begin{table}[t!]
\centering
\caption{Low resource ImageNet models from TIMM github \cite{wightmanTIMM}. Top1/Top5, input image sizes, and parameter count listed. Index corresponds to axis labels in Figure \ref{fig:small_models}}
\label{tab:smallest_models}
\resizebox{\textwidth}{!}{
\begin{tabular}{rlrrrr}
\toprule
 index &                            model &    top1 &    top5 &  img size &  params x$10^6$\\
\midrule
     0 &                  efficientnet\_b2 \cite{tan2019efficientnet} &  80.608 &  95.310 &       288 &         9.11 \\
     1 &                  efficientnet\_b1 \cite{tan2019efficientnet} &  78.792 &  94.342 &       256 &         7.79 \\
     2 &           efficientnet\_b1\_pruned \cite{aflalo2020knapsack} &  78.242 &  93.832 &       240 &         6.33 \\
     3 &       mobilenetv3\_large\_100\_miil \cite{howard2019searching} &  77.914 &  92.914 &       224 &         5.48 \\
     4 &                 mobilenetv2\_120d \cite{sandler2018mobilenetv2} &  77.294 &  93.502 &       224 &         5.83 \\
     5 &            mobilenetv3\_large\_100 \cite{howard2019searching} &  75.768 &  92.540 &       224 &         5.48 \\
     6 &                   mobilenetv3\_rw &  75.628 &  92.708 &       224 &         5.48 \\
     7 &                 mobilenetv2\_110d \cite{sandler2018mobilenetv2} &  75.052 &  92.180 &       224 &         4.52 \\
     8 &             pit\_ti\_distilled\_224 \cite{heo2021rethinking}&  74.536 &  92.096 &       224 &         5.10 \\
     9 &  deit\_tiny\_distilled\_patch16\_224 \cite{touvron2021training} &  74.504 &  91.890 &       224 &         5.91 \\
    10 &                  mobilenetv2\_100 &  72.978 &  91.016 &       224 &         3.50 \\
    11 &                         resnet18 \cite{he2016deep} &  69.758 &  89.078 &       224 &        11.69 \\
\bottomrule
\end{tabular}
}
\end{table}
 For this example, we have selected a set of models from the TIMM model repository \citep{wightmanTIMM} that have between $3.5\times10^6$ and $11.7\times10^6$ parameters. Each model has been pre-trained on the Imagenet dataset \citep{ILSVRC15}. Table~\ref{tab:smallest_models} lists the specific models, their top1/top5 accuracy, image input size, and number of parameters. We have constructed heat maps of the CEV/SDE values by calculating the interaction between each model and building an adjacency matrix. In both Table \ref{tab:smallest_models} and Figure \ref{fig:small_models} we sort the models by Top-1 accuracy. With our constructed matrices we can quickly glance and observe that mobilenetv3\_large100 on row 5 column 5 stands out clearly in the CEV/SDE matrices. We see that although the model has comparable accuracy and parameters to mobilenetv3\_rw and mobilenetv2\_110d, it is actually measured to have worse trade-offs of FPR/FNR w.r.t to the tables best model in terms of accuracy efficientnet\_b2, and is no better or worse than several of the next several models on our accuracy sorted list. CEV and SDE have  prevented us from making a poor selection with relative ease. Again, we find accuracy alone is a poor indicator of model quality.

\subsection{Fairness}
Fairness has rightly been enjoying increased attention over the last several years when measured by the total number of papers addressing it~\cite{caton2020fairness}. Fairness is often defined as the ability of a model to classify all groups within the testing data equally well. For example, a model trained to recognize human faces should be equally good at recognizing the faces regardless of demographic traits (e.g race, gender, age). Unfortunately, unintentionally biased data collected in real-world datasets and even train methodologies can cause undesired performance in models. Our metrics were developed specifically to measure the bias of classifiers, but we will demonstrate they may also be used for measuring fairness as well. Importantly, this methodology allows the metrics to measure fairness in multi-class examples.

To measure bias with CEV or SDE, one model is compared to another. This process can be adapted to measure fairness by comparing a models performance on its test data to its performance on a subset of its test instances. For this purpose, we will select from specific protected attributes and calculate bias with respect to the groups. A large value in CEV or SDE will indicate that per-class bias is increased for one group in the data, and that the model's performance is lower for that group.


\subsubsection{Binary Classification}

\begin{table}[t!]
\centering
\caption{Comparison of Error Rate Equality Difference(ERED)~\cite{dixon2018measuring} metrics, and Difference in Expected Value(DEV)~\cite{hinnefeld2018evaluating} Metrics with our proposed CEV and SDE on measuring bias. We make our comparison to measure fairness traits of models classifying the Titanic dataset \cite{titanic}. CEV/SDE are calculated w.r.t the whole dataset errors, and given protected group. All values are averaged over 5 runs of train/test.}
\label{tab:fair_mets}
\resizebox{.95\columnwidth}{!}{
\begin{tabular}{lrrrrrrrr}
\toprule
\multicolumn{1}{c}{Model} & \multicolumn{4}{c}{Our Metrics} & \multicolumn{4}{c}{Existing Metrics} \\ \midrule
        \multicolumn{1}{l}{-} & \multicolumn{2}{c}{CEV} & \multicolumn{2}{c}{SDE} & \multicolumn{2}{c}{ERED} & \multicolumn{2}{c}{DEV} \\
         - & All→Men & All→Women & All→Men & All→Women & FPED &   FNED & DIMS &   DIAMR \\
\midrule
     NN & 0.013557 &	0.012737 &	0.115002 &	0.093218 & 0.548443 &	0.458016 &	-0.269742 & 0.288790 \\
     SVM & 0.012089 & 	0.000736 &	0.109744 &	0.027081 &	0.412500 & 	0.593508 &	-0.067460 &	0.491071 \\
     GTB & 0.000107 &	0.000941 &	0.010341 &	0.030619 &	0.458462 &	0.513932 &	-0.193700 &	0.364831 \\
\bottomrule
\end{tabular}
}
\end{table}

To demonstrate measuring fairness in binary classification, we trained several common machine learning models on the Titanic dataset~\cite{titanic}: a shallow neural network(NN), a support vector machined(SVM), and a gradient tree boosting classifier(GTB) . This dataset offers information about Titanic passengers with the labels Survived and Did Not Survived. The sex of each passenger is included as a feature of each instance. Sex was excluded in the model training and used later for group-wise fairness testing. These metrics are presented along with the False Positive Equality Difference(FPED), False Negative Equality Difference(FNED)\cite{dixon2018measuring}, Difference in Mean Scores(DIMS), and Difference in Average Model Residuals\cite{hinnefeld2018evaluating} in Table \ref{tab:fair_mets}. 

The four metrics presented for comparison are all zero for perfectly fair predictions. The relatively small value generated for each of the eight metrics is an effect of the small size of the dataset. The fact the FPED, FNED, DIMS, and DIAMR are not 0, shows that some unfairness has been learned by our neural network. The differences in the CEV and SDE scores moving from all data to men only, and all data to women only also indicates biased and unfair performance by the classifier. So we can confirm that in tasks of binary classification, our new metrics conform to the established work in the field of fairness. But as will be shown in Section~\ref{sec:mc fairness}, CEV and SDE are not limited to the analysis of binary labels.


\subsubsection{Multi-Class Classification}
\label{sec:mc fairness}

\begin{figure*}[!h]
    \centering
    \includegraphics[width=\textwidth]{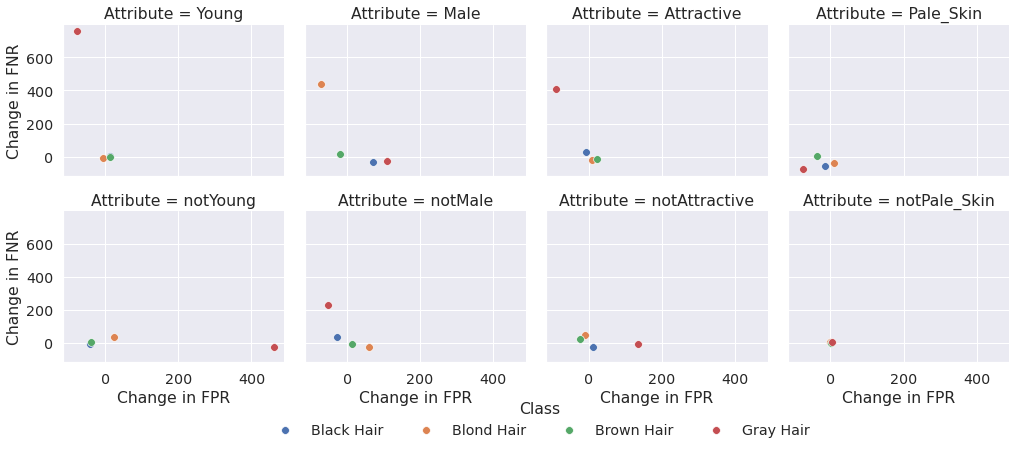}
    \caption{Change in FP/FN rate for protected subgroups of ResNet model trained on CelebA dataset. Change is calculated w.r.t the to complete validation set}
    \label{fig:group_fairness}
\end{figure*}

We have asserted that CEV and SDE can be used to measure fairness in multi-class classification. We will now demonstrate that process using the CelebA dataset~\cite{liu2015faceattributes}. This dataset contains several thousand images of celebrates and public figures with 40 binary attributes. We have selected from the provided attributes a subset representing hair color to serve as training labels. We then trained a ResNet34 image recognition model to identify the hair color of the image subjects. From the remaining provided attributes, we have identified several to serve as protected groups (``Attractive'', ``Male'', ``Pale Skin'', ``Young''). As these labels come from what might be described as ``privileged'', we also consider subsets formed from the conjugate of these labels. We should note that the conjugate does not imply the opposite. The absence of a Pale Skin label for example does not explicitly mean dark skin but would contain all of those examples. 

The results are contained in Figure~\ref{fig:group_fairness} and Table~\ref{celebA_FNR}. We find that groups ``Male'' and ``Pale Skin'' have the highest Top-1 Accuracy. However, we also find they have high levels of class unfairness. Specifically for Male, our model is far less likely to correctly identify Male as having Blond Hair, and more likely to incorrectly guess they have Gray Hair. Meanwhile, ``Not Pale Skin'' has lower accuracy, but the accuracy and FPR/FNRs are much closer to the average of the model as a whole. This is easily visible in Figure \ref{fig:group_fairness}. This unevenness is neatly captured by the corresponding CEV and SDE values or the groups in our data.

\begin{table}
\centering
\caption{Top-1, CEV, SDE, and change in FPR/FNR for selected protected class from ResNet model trained on CelebA dataset}
\resizebox{\columnwidth}{!}{
\label{celebA_FNR}
\begin{tabular}{lrrrrr}
\toprule
{Protected Attribute } &  Top-1 &  CEV &  SDE &  Change in FPR &  Change in FNR \\
\midrule
Full Test Set           &    0.9212 &         {} &         {} &         {} &         {} \\ 
Attractive    &    0.9222 &         0.0015 &         0.0331 &       -31.0809 &        80.4380 \\
Male          &    0.9225 &         0.1413 &         0.2205 &        12.8440 &        77.8003 \\
Pale Skin     &    0.9224 &         0.0035 &         0.0465 &       -43.8572 &       -33.9335 \\
Young         &    0.9215 &         0.0002 &         0.0082 &       -27.6765 &       150.5065 \\
Not Attractive &    0.9208 &         0.0034 &         0.0493 &        45.6423 &         6.8297 \\
Not Male       &    0.9207 &         0.0053 &         0.0562 &         1.2762 &        47.2981 \\
Not Pale\_Skin  &    0.9207 &         0.0000 &         0.0021 &         1.9565 &         1.4648 \\
Not Young      &    0.9213 &         0.0035 &         0.0313 &       146.3057 &         0.2381 \\
\bottomrule
\end{tabular}
}
\end{table}

\section{Discussion and Limitations}
\label{sec:discuss}


As with any metric, it is also important to remember that CEV and SDE are only meaningful in context. A higher value for CEV indicates that the second model has a higher class-wise bias. A higher value for SDE indicates that the second model is skewing towards false positives or false negatives. Either behavior represents a degraded real-world performance for a model in a way that may not be captured by accuracy or precision as demonstrated in Section~\ref{sec:cases}.


The importance of measuring fairness and bias in multi-class data should not be underestimated. Data that meaningfully describes the real world is often multi-class. While it is true to that multi-class classification can be re-framed as many binary classification problems, re-framing a problem as 100 or 1,000 one-vs-each problems would only serve to make reasoning about the implications much more difficult. We believe CEV and SDE are applicable to many real-world problems completely ignored by their binary cousins. 


CEV and SDE can be used to measure the fairness of a machine learning model, but only group fairness. Individual fairness, which is defined as the degree to which similar individuals are classified similarly, is not measured in any of the use cases presented in Section~\ref{sec:cases}.

We have not found any consistent threshold that indicates by itself that a model is or is not biased. It may be that such a threshold exists. Also important to remember that biased performance may be the result of algorithmic bias, or it may be a reflection of biased data and CEV/SDE alone cannot determine its source. But with these limitations in mind, CEV and SDE reliably indicate that one model is more or less biased than another. As concluded in~\cite{hinnefeld2018evaluating}, ''...fairness metrics in machine learning must be interpreted with a healthy dose of human judgment."

CEV and SDE are calculated w.r.t to some other classifier and only classifiers. As such they are not suitable for every situation. However, we believe they provide a good starting point for the community to begin to address measuring more sophisticated machine learning tasks. Additionally, we endeavor to extend the concepts of CEV/SDE to other tasks like image segmentation which are harder still to quantify. We also believe our insights from CEV/SDE can be used to create stand-alone metrics to measure bias and fairness without making direct model comparisons. 

\section{Conclusion}
\label{sec:conclusion}
Unfairness is a persistent and difficult problem in machine learning. Bias is more quantifiable but just as dangerous to the reliable performance of machine learning models in the real world. In this paper, we have introduced two new metrics: CEV and SDE. These metrics can reliably reveal that a model is more or less biased compared to another model. We have also demonstrated that these new metrics can be used to measure the fairness of a model used for classification. Importantly, these metrics are meaningful when used with multi-class data, even with a very large number of classes. 



\bibliography{iclr2022_conference}
\bibliographystyle{iclr2022_conference}


\end{document}

%% file: math_commands.tex
\usepackage{amsmath,amsfonts,bm}

\def\eqref#1{equation~\ref{#1}}

\def\1{\bm{1}}

\DeclareMathAlphabet{\mathsfit}{\encodingdefault}{\sfdefault}{m}{sl}
\SetMathAlphabet{\mathsfit}{bold}{\encodingdefault}{\sfdefault}{bx}{n}

